\documentclass{ecai} 

\usepackage{latexsym}
\usepackage{amssymb}
\usepackage{amsmath}
\usepackage{amsthm}
\usepackage{booktabs}
\usepackage{enumitem}
\usepackage{graphicx}
\usepackage{color}
\usepackage{multirow}
\usepackage{algorithm}
\usepackage{algpseudocode}

\newcommand{\BibTeX}{B\kern-.05em{\sc i\kern-.025em b}\kern-.08em\TeX}
\algrenewcommand\algorithmicrequire{\textbf{Input:}}
\algrenewcommand\algorithmicensure{\textbf{Output:}}

\begin{document}


\begin{frontmatter}





\title{PQ-GCN: Enhancing Text Graph Question Classification with Phrase Features}


\author[A]{\fnms{Junyoung}~\snm{Lee}\thanks{Corresponding Author. Email: junyounglee.k@gmail.com.}\footnote{Ideation of this work was carried out while author was a student of Nanyang Technological University.}}
\author[B]{\fnms{Ninad}~\snm{Dixit}\footnote{Equal contribution.}}
\author[B]{\fnms{Kaustav}~\snm{Chakrabarti}\footnotemark}
\author[B]{\fnms{S.}~\snm{Supraja}}

\address[A]{Independent Reseacher}
\address[B]{Nanyang Technological University}


\begin{abstract}
Effective question classification is crucial for AI-driven educational tools, enabling adaptive learning systems to categorize questions by skill area, difficulty level, and competence. It not only supports educational diagnostics and analytics but also enhances complex downstream tasks like information retrieval and question answering by associating questions with relevant categories. Traditional methods, often based on word embeddings and conventional classifiers, struggle to capture the nuanced relationships in question statements, leading to suboptimal performance. We propose a novel approach leveraging graph convolutional networks, named Phrase Question-Graph Convolutional Network (PQ-GCN). Through PQ-GCN, we evaluate the incorporation of phrase-based features to enhance classification performance on question datasets of various domains and characteristics. The proposed method, augmented with phrase-based features, outperform baseline graph-based methods in low-resource settings, and performs competitively against language model-based methods with a fraction of their parameter size. Our findings offer a possible solution for more context-aware, parameter-efficient question classification, bridging the gap between graph neural network research and its educational applications.
\end{abstract}

\end{frontmatter}





\section{Introduction}
Question classification is vital in providing AI-driven education and training, assisting adaptive learning systems by categorizing questions into skill area or level of difficulty and competence, and providing educational diagnosis and analytics. Accurate automatic question classification also provides headway to more complex tasks such as information retrieval and question answering, in which the association of the given question with a certain category allows for retrieval of relevant contextual knowledge to formulate compatible answers. 

Traditional methods, relying on word embeddings and conventional classifiers, often struggle with capturing the nuanced relationships between words and phrases, leading to suboptimal classification performance. The complexity of natural language, especially in the context of educational and domain-specific questions, requires more sophisticated approaches that can understand and leverage the inherent structure of language.

Aside from text embeddings, graphs have recently been highlighted as a way to represent unstructured text data. Questions can be naturally represented as graphs, where nodes correspond to different text features, and edges capture the relationships between them, such as syntactic dependencies, semantic similarities, or proximity measures. This graph-based representation aligns perfectly with the strengths of Graph Convolutional Networks (GCNs), which excels at learning from structured data. GCNs are designed to operate on graph-structured data, making them ideal for processing and understanding the interconnected nature of text features within questions. This capability enables GCNs to capture the underlying structure and dependencies in a question, leading to more accurate and context-aware classification decisions. GCNs offer a promising solution by modeling the relationships between text features and their syntactic and semantic connections as graphs, which allows for a more nuanced and context-aware classification.

Question classification poses a unique problem even among the different types of text classification, as the amount of information available in a question tends to be limited compared to document-level text counterparts. While neither tackling the task of text classification with GCNs, nor question classification via automated systems is new, our investigation aims to determine the feasibility of adapting GCNs for question classification. Up till our work, GCNs have not been thoroughly explored in the context of question classification, with the exception of \cite{supraja-khong-2024} and \cite{pawestri-question-gat-2024} which are reviewed in Section~\ref{ssec:aqc}. In particular, we explore additional phrase-based features in our proposed model, PQ-GCN, to enhance classification performance, and assess the effectiveness of these features in low-resource settings, where there is limited labeled data available.

\section{Related Work} \label{sec:related}

\subsection{Automatic Question Classification} \label{ssec:aqc}

One of the earliest works on question classification was in~\cite{li-roth-2002-learning} which made use of Support Vector Machines (SVMs) and explored a variety of features like lexical, synctactic, and semantic features. Another reference work by~\cite{huang-etal-2008-question} also discusses usage of head words and hypernyms as features for SVM and Maximum Entropy (ME) algorithms. They demonstrated the effectiveness of statistical learning methods in classifying questions, while rule-based classification~\cite{hovy-etal-2002-typology} via string matching in questions have also been explored in earlier works.

\begin{figure*}[t]
\centering
\includegraphics[width=0.8\textwidth]{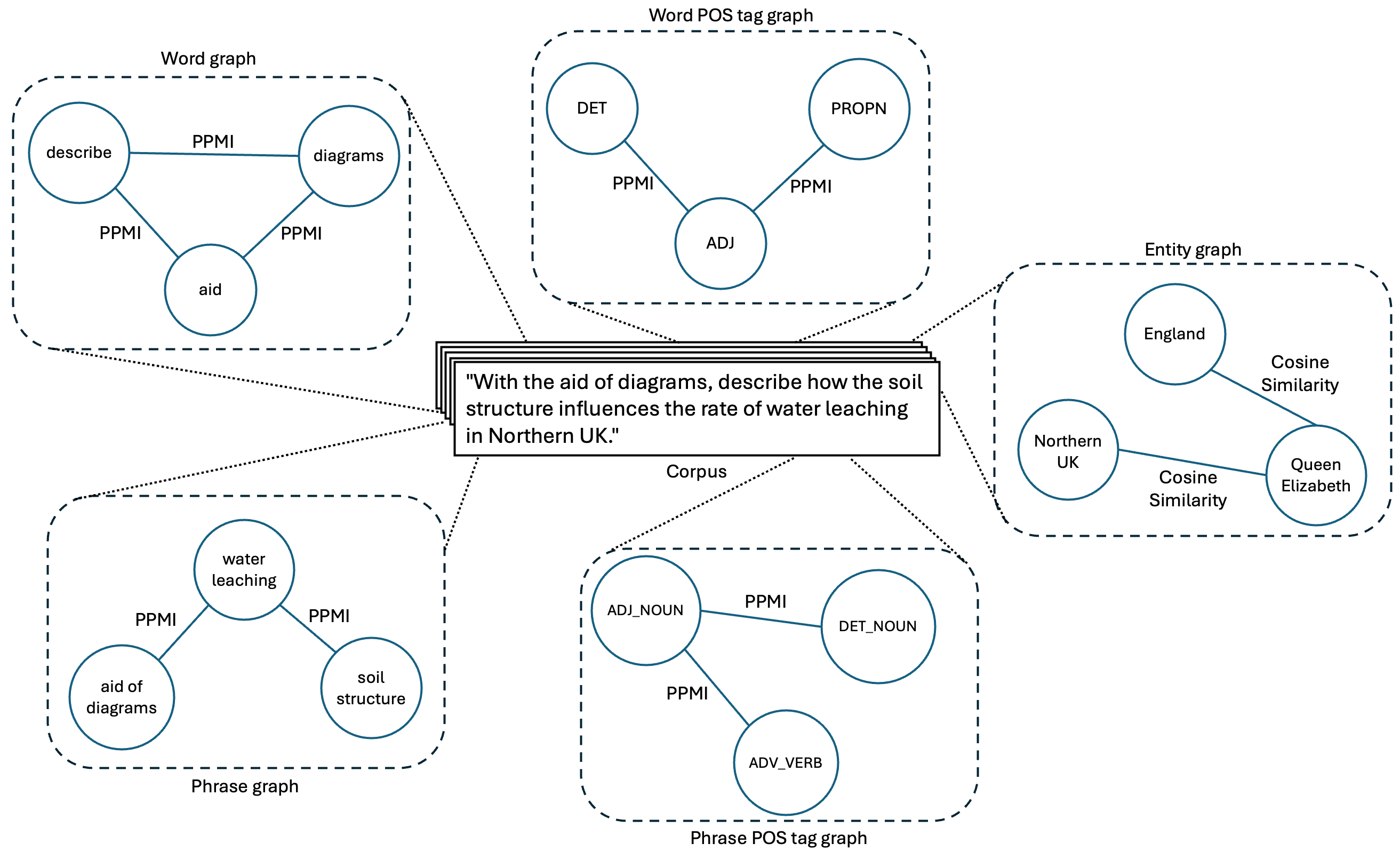}
\caption{Corpus-level graph construction with extracted features. The various features extracted from each question are consolidated at the corpus level to form each feature graph.}
\label{fig:construction}
\end{figure*}

Following these works, a review of automatic question classifiers from 2012 to 2017~\cite{silva-etal-2019-systematic} revealed that there was an increase in the number of works proposing neural network-based methods, or using neural networks in combination with rule-based methods. The approaches include concatenating pre-trained word embeddings to form a question embedding~\cite{godea-nielsen-2018-annotating}, incorporating subword information in character-based Convolutional Neural Network (CNN)~\cite{pota2018ijcnn}, and combining CNN and Long Short-Term Memory (LSTM) in a hybrid model~\cite{Zhou2016/12}. These neural networks, effectively balancing feature extraction and sequence modeling, were shown to outperform traditional machine learning techniques and earlier deep learning models in classifying questions across various domains. More recently, Pre-trained Language Models (PLMs) and Large Language Models (LLMs) which have been trained on large amounts of text data have shown their success in natural language understanding-based tasks such as question generation and classification~\cite{ALFARABY2024100298}), but questions might have domain-specific terminologies or technical jargon which may not be captured fully by such generalized models.

While there are other works which focus on question classification using neural networks, we review the two works up to date that specifically makes use of text graphs and highlight the improvements we propose with PQ-GCN. The first work~\cite{supraja-khong-2024} incorporates dependency parsing between words, Pointwise Mutual Information (PMI) between regular expressions, cosine similarity between phrase embeddings, and KL divergence between topics in a heterogeneous graph for question and short document classification. The authors based the architecture on TensorGCN~\cite{Liu_You_Zhang_Wu_Lv_2020} and 4 heterogeneous node sets and adjacency matrices, while PQ-GCN is based on SHINE~\cite{wang-etal-2021-hierarchical} and uses 5 homogeneous node sets and adjacency matrices which reduces sparsity in input data. The second work~\cite{pawestri-question-gat-2024} uses word graph based on dependency parsing, and trains a Graph Attention Network (GAT)~\cite{veličković2018graph}. This approach lacks consideration for other textual features, such as semantic and sequential relationships between words and phrases, that would enable better understanding of questions.

\subsection{Text Classification Using Text Graphs}

TextGCN~\cite{Yao_Mao_Luo_2019} pioneered approaching text classification as a node classification problem. Yao et al. constructs a corpus-level graph with document nodes and word nodes---PMI as word-word edge weights and Text Frequency-Inverse Document Frequency (TF-IDF) as word-document edge weights---and applies a 2-layer GCN to obtain the class output. Several works propose similar graph construction and node initialization based on TextGCN, with more efficient or better regularized graph propagation~\cite{pmlr-v97-wu19e-sgc,zhu2021s2gc}.

Other works have also explored multi-graphs instead of a single corpus-level graph. TensorGCN~\cite{Liu_You_Zhang_Wu_Lv_2020} builds a tensor graph from three separate graphs: semantic graph via word embeddings, syntactic graph via grammar parsing, and sequential graph via PMI values. Each of these graphs has different word-word edge weights while sharing the same TF-IDF values for word-document edge weights, and they undergo intra- and inter-graph propagation. ME-GCN~\cite{wang2022megcnmultidimensionaledgeembeddedgraph} constructs a graph that has multi-dimensional word-word, word-document, and document-document edges, using Word2Vec and Doc2Vec embeddings for node features, and normalized embedding distance and TF-IDF for edge features. SHINE~\cite{wang-etal-2021-hierarchical} uses word graphs based on three word-level relationships---PMI, POS tag co-occurrence, and cosine similarities between embeddings of entity pairs appearing in the corpus---which are easy to obtain at a low cost using common natural language processing techniques.

While node classification is a transductive approach, where one needs to build a corpus-level graph consisting both train and test document nodes, InducT-GCN~\cite{inductgcn} extends the task into a inductive approach, where a training corpus graph is constructed first similarly to TextGCN, and any new test corpus is built as a virtual subgraph for inference.

Graph classification, in which you classify a document graph as opposed to a document node, has also been used as an inductive approach to text classification. Text-Level-GNN~\cite{huang-etal-2019-text} uses a sliding window to build multiple graphs, each with a small number of nodes and edges, with trainable edge weights that are shared across the graphs for the same word pair. Huang et al. also uses a message passing mechanism instead of a GCN architecture, relying on aggregating neighborhood information via max-pooling and combining them via weighted sums, instead of convolution functions. 
Graph Fusion Network~\cite{DAI2022107659} constructs four individual global word co-occurrence graphs, and each document is represented as with a set of four subgraphs. GCN is applied to each subgraph, and the resulting output is concatenated, passed through an MLP layer, and average-pooled to obtain the final document embedding.

There are also combinatory methods based on GAT. HyperGAT (Hypergraph Attention Networks)~\cite{ding-etal-2020-less} builds hypergraphs for each document to capture high-level interaction between words, consisting of sequential hyperedges connecting all words in a sentence and semantic hyperedges connecting top-K words with LDA-based topic modeling. Attention mechanism is utilized at both node- and edge-level to update representations. 

While these graph-based methods focus on general document-level text classification, we identify a research gap in adapting them for question classification, with the lack of question-specific feature extraction and graph construction methods, given questions' comparatively shorter lengths and interrogative sentence structure.


\section{Why Use Phrase Features?}\label{sec:whyphrase}

As with all text classification tasks, feature extraction from questions poses a particular challenge, as the features need to be representative of the questions' characteristics and yet be domain-agnostic to be generalizable across different classification frameworks. 

There is growing agreement in literature that our mental lexicon contains formulaic language, including idioms, phrases, and multi-word expressions~\cite{Conklin_Schmitt_2012}, and there are positive benefits from understanding text through parsing by larger, more meaningful chunks rather than individual words~\cite{Abney1992}. There is sufficient linguistic basis to explore beyond word-level representations in improving natural language understanding capabilities of automatic systems. While works discussed in the previous section discuss word and document relationships, phrase relationships have not been explored in the context of text classification with graph representations. Li and Roth~\cite{li-roth-2002-learning} discuss the use of \textit{non-overlapping chunks}, extracted via a trained classifier, but not explicitly defined phrases. 

Parsing by phrases and extracting phrase features have already been proven effective in some areas of computational linguistics, specifically statistical machine translation~\cite{koehn-etal-2003-statistical}, neural machine translation~\cite{wang-etal-2017-translating}, and topic modeling for questions~\cite{supraja-etal-2021-regularized,supraja-khong-2024}.

While word-level features such as embeddings and POS tags are foundational for understanding the basic structure and meaning of a question, phrase-level features offer a more comprehensive approach to understanding the question text, that can capture meaningful, disambiguated chunks of information. By extracting noun phrases and verb phrases and analyzing their semantic embeddings and relationships, one can capture the nuanced meaning and intent of questions, leading to more accurate and effective classification models.

\section{Methodology}

\subsection{Feature Extraction and Graph Construction} \label{ssec:features}

Given a corpus of question-label pair, we perform basic text cleaning which includes removal of contractions and punctuation. 

To construct meaningful graph representations from a corpus of question text, we select several feature extraction techniques, each contributing uniquely to capturing different aspects of the textual data. A summary of the extracted features are provided in Table~\ref{tab:extracted-features}.

\begin{table}[t!]
    \centering
    \caption{Summary of Extracted Features. Node features for Word POS and Phrase POS are obtained via one-hot encoding, as no viable pre-trained embeddings are available for POS tags.}
    \label{tab:extracted-features}
    \begin{tabular}{lccc}
        \toprule
        & \textbf{Node Type} & \textbf{Node Feature} & \textbf{Edge Weight}  \\
        \midrule
        \textbf{Word PPMI}    & Word & Word2Vec & PMI value  \\
        \textbf{Word POS} & POS tag & - & PMI value   \\
        \textbf{Phrase PPMI}    & Phrase & PhraseBERT & PMI value  \\
        \textbf{Phrase POS}    & Phrase POS tag & - & PMI value \\
        \textbf{Named Entities}    & Named entity & TransE & Cosine similarity \\
        \bottomrule
    \end{tabular}
    
\end{table}

\begin{figure*}[t]
\centering
\includegraphics[width=0.8\textwidth]{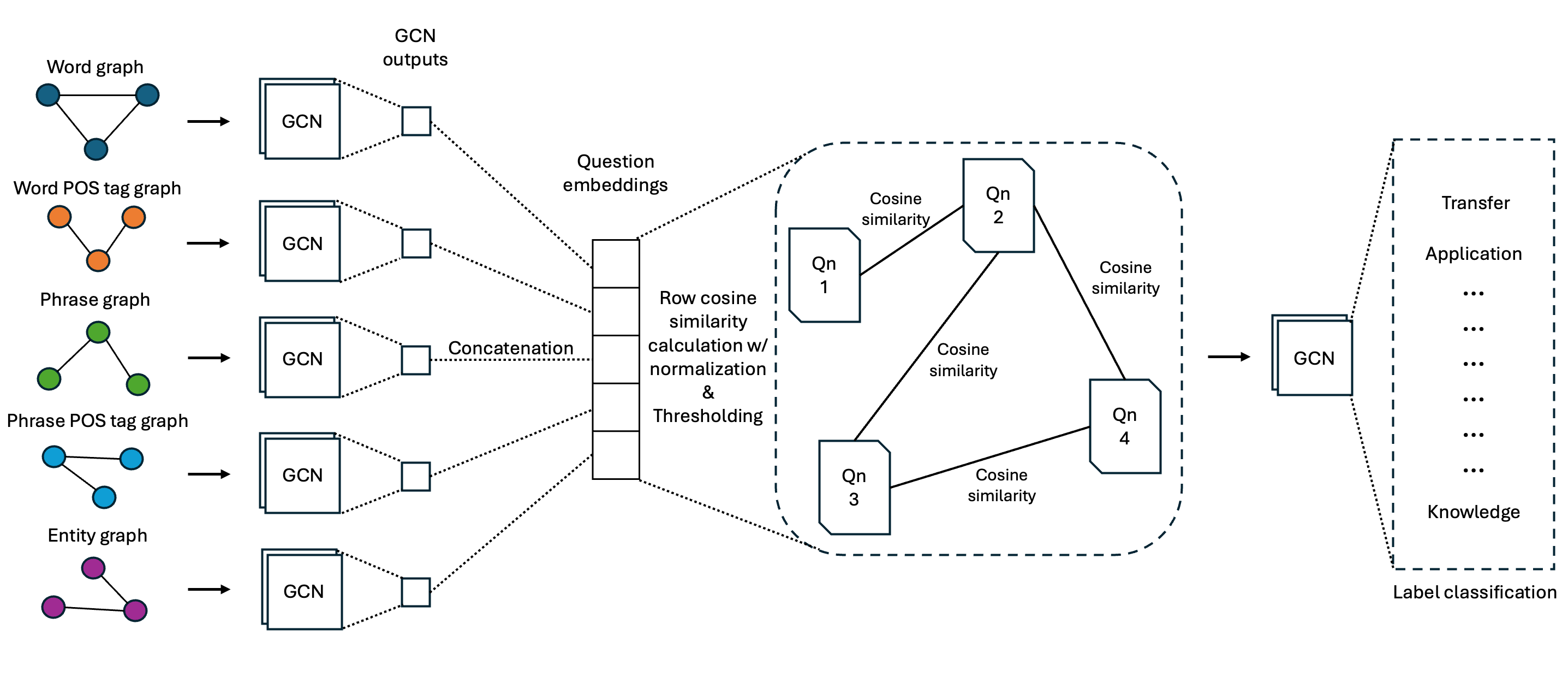}
\caption{Overview of model architecture. Each graph is passed through their own 2-layer GCN. Each GCN output is then concatenated to form question embeddings, and a question graph is created by using these question embeddings as node features and calculating the cosine similarity between the question embeddings for edge features. Hence, a dynamic graph of question nodes and cosine similarity edges is learned from the individual 2-layer GCN outputs, and this dynamic graph is propagated through a final 2-layer GCN and a linear layer for label classification.}
\label{fig:propagation}
\end{figure*}

\paragraph{Words and PMI}
We first tokenize the given question corpus by whitespace to obtain a unique set of words. A word-level graph is constructed, with word nodes connected by edges weighted with PMI values, representing local co-occurrence relationships. Specifically, we utilize positive PMI (PPMI) with a window size of 5, given by max(PMI($word_1, word_2$), 0). PMI is given by:

\begin{equation}
    \text{PMI}(w_1, w_2) = \log \frac{P(w_1, w_2)}{P(w_1)P(w_2)}
\end{equation}

where $P(w_1) = \frac{\#W(w_1)}{\#W}$, $P(w_2) = \frac{\#W(w_2)}{\#W}$, $P(w_1, w_2) = \frac{\#W(w_1, w_2)}{\#W}$. $\#W(w)$ is the number of sliding windows that a word $w$ in the corpus, and $\#W(w_1, w_2)$ is the number of sliding windows that words $w_1$ and $w_2$ appear together in the corpus. $\#W$ is the total number of sliding windows in the corpus.

Pre-trained word embeddings from word2vec~\cite{mikolov2013efficientestimationwordrepresentations} are also obtained for each word in the corpus, to be used as additional semantic information by concatenating with node embeddings.

\begin{table}[t!]
    \centering
    \caption{PQ-GCN Model Parameters. Total Parameter Size varies according to number of label classes and total number of unique word-level and phrase-level POS tags (which determines embedding size via one-hot encoding), which differ in each dataset. We report the \underline{maximum} parameter size of the model that is evaluated on TREC dataset.}
    \label{tab:pq-gcn_parameters}
    \begin{tabular}{@{}lc@{}}
    \toprule
    
    Word Embedding Size      & 300              \\
    Phrase Embedding Size      & 768              \\
    Entity Embedding Size      & 100              \\
    Sliding Window Size for PMI & 5              \\
    Hidden Dimension Size   & 200             \\
    No. of Layers in GCNs   & 2             \\
    Dropout & 0.7 \\
    Threshold for Question Cosine Similarity & 2.7 \\
    \midrule
    Optimizer & Adam \\
    Learning Rate & 1e-3 \\
    Weight Decay & 1e-4 \\

    \midrule
    Total Parameter Size & 495,406 \\
    
    \bottomrule
    \end{tabular}
\end{table}

\paragraph{Word-Level POS Tags}
POS tags for each word is obtained via the default set from NLTK~\footnote{\url{https://nltk.org}} and PPMI values between each pair of POS tags are calculated as edge weights for a POS tag graph.

\paragraph{Phrases and PMI}
Phrases are extracted from each question text via POS tag regex matching for noun and verb phrases with \texttt{spaCy}~\footnote{\url{https://spacy.io}} in Python, and a set of unique phrases in the corpus is obtained. The regex patterns for noun and verb phrases are given by (\ref{eq:np}) and (\ref{eq:vp}) respectively:

\begin{align}
    \mu_{N} = &\langle\text{DET}\rangle?\langle\text{NUM}\rangle* \nonumber \\
    &(\langle\text{ADJ}\rangle\langle\text{PUNCT}\rangle?\langle\text{CONJ}\rangle?)* \nonumber\\
    &(\langle\text{NOUN}\rangle|\langle\text{PROPN}\rangle\langle\text{PART}\rangle?)+\label{eq:np}\\
    \mu_{V} = &\langle\text{AUX}\rangle*\langle\text{ADV}\rangle*\langle\text{VERB}\rangle\label{eq:vp}
\end{align}

\noindent where $?$, $*$, and $+$ represent zero or one, zero or more, and one or more occurrences of the preceding POS tag respectively.

Each question is represented as a sequence of verb phrases and noun phrases present in the text. Then a phrase-level graph (similar to the word-level graph) is constructed with PMI-weighted edges. Phrase embeddings are obtained from PhraseBERT~\cite{wang-etal-2021-phrase} to provide additional semantic information.

    


\paragraph{Phrase-Level POS Tags}
During the phrase extraction process above, \texttt{spaCy} also provides word-level POS tags for each word in the extracted phrases. These word-level POS tags are concatenated to form phrase-level POS tags. For example, in the question ``Discuss the main objective of layout design rules",  using the regex $\langle\text{ADJ}\rangle\langle\text{NOUN}\rangle$, a noun phrase [main objective] is extracted, together with the corresponding phrase-level POS tag of $\langle\text{ADJ\_NOUN}\rangle$. Then, the PPMI values between each pair of phrase-level POS tags are calculated as edge weights for a phrase-level POS tag graph.  

\paragraph{Named Entity Recognition (NER)}
Named entities (e.g., people, organizations, locations) carry significant meaning and often denote key elements within a question. A list of named entities is obtained from the NELL knowledge base~\cite{NELLKB}, and is used to extract named entities via string matching. With the short text nature of questions, it is difficult to obtain multiple entities from a single question to calculate co-occurrence statistics. Hence, we utilize TransE embeddings~\cite{NIPS2013_TransE} for each named entity and calculate the cosine similarity between embeddings of each pair of named entities, to be used as edge weights in constructing a named entity graph.

From the corpus, the above features are extracted and the respective relationships are computed to obtain edge weights, which are then used to construct the graph for each feature. The examples of graphs constructed are shown in Figure~\ref{fig:construction}. For the corpus-level graphs constructed above, the node features are initialized as one-hot vectors except for entity graph, which uses the TransE embeddings directly as node features. The mapping of which text features are present in which question is used to generate question-level embeddings during propagation. By combining these diverse feature extraction methods, we construct a comprehensive and multi-faceted graph representation of the question text, capturing various semantic, syntactic, and named entity information crucial for question classification.

\subsection{Model Architecture}

As we develop phrase-based features to bridge the gap towards building a more robust automatic question classification model, we also identify a model architecture capable of incorporating phrase-based features as a modular add-on, as well as consolidating each graph of different node and edge types into a single question embedding. Initial approach was to concatenate the outputs of 5 individual 2-layer GCNs, followed by a linear layer and softmax function. Various combinations of 2D convolution layers and pooling layers have also been experimented with, but ultimately we found that the final question embedding was not able to provide sufficient information for classification and performed worse than baseline models.

Following the initial exploration, we base our output layers largely on SHINE~\cite{wang-etal-2021-hierarchical}. An overview of model architecture is provided in Figure~\ref{fig:propagation}, and after a parameter sweep across different layer types, we provide the model and training parameters for the proposed PQ-GCN in Table~\ref{tab:pq-gcn_parameters}. A training iteration is described in Algorithm~\ref{alg}

\begin{algorithm}
\caption{Pass-through Algorithm for PQ-GCN}\label{alg}
\begin{algorithmic}[1]
\Require $\{(V^{(i)}, A^{(i)})\}_{i=1}^{5}$: Node sets and adjacency matrices for constructed graphs
\Require $\text{GCN}_i$: Two-layer GCN for the $i$-th graph
\Require $\text{GCN}_Q$: Two-layer GCN for question-level graph
\Require $f_{\text{linear}}$: Final linear layer for classification
\Ensure Predicted labels for each question

\For{$i = 1$ to $5$}
    \State $H^{(i)} \gets \text{GCN}_i(V^{(i)}, A^{(i)})$ 
\EndFor

\State $H_Q \gets \text{Concat}(H^{(1)}, H^{(2)}, \dots, H^{(5)})$ 

\State $A_Q \gets \text{CosineSimilarity}(H_Q, H_Q)$ 

\State $Z_Q \gets \text{GCN}_Q(H_Q, A_Q)$ 

\State $\hat{Y} \gets f_{\text{linear}}(Z_Q)$ 

\end{algorithmic}
\end{algorithm}

The intermediate states $H^{(i)}$, $H_Q$, $A_Q$, and $Z_Q$ refer to GCN output for graph $i$, question embeddings, question-level dynamic adjacency matrix, and GCN output for question-level graph respectively.

\begin{table*}[t!]
    \centering
    \caption{Classification label distributions across evaluation datasets}
    \label{tab:label_distributions}
    \begin{tabular}{@{}lrrrrr@{}}
    \toprule
    {\textbf{Dataset}}& 
                             \textbf{NU} & \textbf{ARC} & \textbf{LREC} & \textbf{Bloom} & \textbf{TREC}  \\
    \midrule
    \textbf{Dataset Size}& 596 & 279 & 344 & 2522 & 5952 \\
    \textbf{Classification} & Cognitive complexities & Reasoning capabilities & Expected answer types & Educational objectives  & Question topics   \\
    \textbf{Avg. \# of Words} & 10.8 & 24.8 & 16.1 & 14.8  & 10.0   \\
    \midrule
    \multirow{7}{*}{\textbf{Class Labels}} & Transfer (49.7\%) & {Linguistic matching} & Very short answer (41.6\%) & Comprehension (38.1\%)  & Entity (22.6\%)   \\
    & Knowledge (33.5\%) & (47.0\%) &Context sensitive (36.0\%) & Knowledge (13.6\%) & Description and abstract \\
    & Application (16.8\%)& Basic facts (28.0\%) & Answers will vary (22.4\%)&Evaluation (12.6\%) & concept (21.8\%)\\
    &  & Hypothetical (25.0\%) & &Application (12.5\%) & Human being (21.6\%)\\
    & & & &Analysis (12.1\%) & Numeric value (17.0\%)\\
    & & & &Synthesis (11.1\%) & Location (15.4\%)\\
    & & & & & Abbreviation (1.6\%)\\
    
    \bottomrule
    \end{tabular}
\end{table*}

\section{Experiments}

\subsection{Datasets}

We first select a number of datasets used for question classification task to evaluate our proposed model on.

\paragraph*{NU} This dataset was obtained from Najran University~\cite{YAHYA2013587}, containing 596 questions with 3 class labels of different cognitive complexities. 

\paragraph*{ARC} This dataset was released as part of the A12 Reasoning Challenge~\cite{boratko-etal-2018-systematic}, consisting 279 questions labeled with 3 different reasoning capabilities.

\paragraph*{LREC} This is a subset of 344 questions from the original dataset of science questions asked by teachers in real middle school classrooms, with a proposed taxonomy by expected answer types~\cite{godea-nielsen-2018-annotating}. The subset was chosen by selecting 3 mutually exclusive class labels---'Very short answer', 'Context sensitive', and 'Answers will vary'.

\paragraph*{Bloom} Bloom's Taxonomy~\cite{bloom1956taxonomy} is one of the most well-established question classification frameworks used by educators, based on different educational objectives. This dataset is a set of questions collected from different web sources and manually classified into the six cognitive levels of Bloom's Taxonomy\footnote{The dataset can be found at \url{dx.doi.org/10.13140/RG.2.1.4932.3123}}.

\paragraph*{TREC} The Text REtrieval Conference (TREC) dataset contains $\sim$6000 questions from four sources---4,500 English questions published by USC~\cite{hovy-etal-2001-toward}, about 500 manually constructed questions for a few rare classes, 894 TREC 8 and TREC 9 questions, and also 500 questions from TREC 10 which serves as the test set. These questions were manually labeled into 6 class labels.

Aside from TREC, for which the dataset split is obtained from HuggingFace~\footnote{\url{https://huggingface.co/datasets/CogComp/trec}}, the train/test splits are determined by a fixed seed and can be provided upon request. Relevant statistics for the datasets are found in Table~\ref{tab:label_distributions}.

\subsection{Baseline Models}

We then compare the performance of our proposed model against selected baseline models in Table~\ref{tab:performance-comparison}.  We include vanilla CNN and Bi-LSTM (with parameters in Table~\ref{tab:cnn_parameters} and Table~\ref{tab:bilstm_parameters}) for non-graph classification models; fine-tuned BERT (\texttt{bert-base-uncased}\footnote{\url{https://huggingface.co/google-bert/bert-base-uncased}}) for pre-trained language model. The baseline graph models were selected by reviewing their contributions in graph neural networks for text classification~\cite{Wang2024}, each using distinct feature engineering and graph construction methods for text data, as in Section~\ref{sec:related}, as compared to works which have proposed architectural improvements for computational efficiency. With the exception of CNN, Bi-LSTM, and BERT, the rest of the baseline models have their model and training parameters set as provided by original authors in their project repositories.

\begin{table}[t!]
    \centering
    \caption{CNN Model Parameters}
    \label{tab:cnn_parameters}
    \begin{tabular}{@{}lcccc@{}}
    \toprule
    \textbf{Parameter} & \textbf{Layer 1} & \textbf{Layer 2} & \textbf{Layer 3} & \textbf{Layer 4} \\
    \midrule
    Type                & Conv2d          & Conv2d           & Conv2d           & Linear \\
    Embedding Size      & 300             & --               & --               & --               \\
    Sequence Length     & 15             & --               & --               & --               \\
    Number of Filters   & 100              & 100              & 100              & --              \\
    Kernel Size         & 3x3             & 4x4              & 5x5              & --              \\
    Stride             & 1x1             & 1x1              & 1x1              & --              \\
    Activation Function & --            & --             & --             & Softmax          \\
    Dropout Rate        & --            & --             & --           & 0.5               \\
    \bottomrule
    \end{tabular}
\end{table}

\begin{table}[t!]
    \centering
    \caption{Bi-LSTM Model Parameters}
    \label{tab:bilstm_parameters}
    \begin{tabular}{@{}lcccc@{}}
    \toprule
    \textbf{Parameter} & \textbf{Layer 1} & \textbf{Layer 2} & \textbf{Layer 3}  \\
    \midrule
    Type                & Bi-LSTM         & Bi-LSTM             & Linear \\
    Embedding Size      & 300             & --               & --               \\
    Hidden Dimension Size   & 128              & 128              &    --       \\
    Activation Function & --            & --             & Softmax              \\
    \bottomrule
    \end{tabular}
\end{table}

\begin{table*}[t!]
    \centering
    \caption{Baseline comparisons for macro-averaged F1, Precision, and Recall scores across datasets. Best F1 scores are in \textbf{bold} and the second best F1 scores are \underline{underlined} for each dataset.}
    \label{tab:performance-comparison}
    \begin{tabular}{lllccccc}
        \toprule
        \textbf{Type} & \textbf{Model} & \textbf{Metric} & \textbf{NU} & \textbf{ARC} & \textbf{LREC} & \textbf{Bloom} & \textbf{TREC}  \\
        \midrule
        \multirow{6}{*}{Non-graph models} & \multirow{3}{*}{CNN} & F1        & 0.085 & 0.226 & 0.198 & \underline{0.752} & 0.782 \\
                                 && Precision & 0.048 & 0.171 & 0.141 & 0.763 & 0.783  \\
                                 && Recall    & 0.333 & 0.333 & 0.333 & 0.750 & 0.790  \\
        \cmidrule{2-8}
        & \multirow{3}{*}{Bi-LSTM} & F1        & 0.607 & 0.564 & 0.480 & 0.425 & 0.653  \\
                                 && Precision & 0.606 & 0.554 & 0.511 & 0.465 & 0.692  \\
                                 && Recall    & 0.620 & 0.598 & 0.495 & 0.422 & 0.642  \\
        \midrule
        \multirow{3}{*}{Pre-trained language model} & \multirow{3}{*}{BERT} & F1        & 0.703 & 0.674 & 0.553 & \textbf{0.829} & \textbf{0.970} \\
                                 && Precision & 0.828 & 0.777 & 0.763 & 0.846 & 0.981  \\
                                 && Recall    & 0.680 & 0.717 & 0.597 & 0.818 & 0.961  \\
        \midrule
        \multirow{24}{*}{Graph models} & \multirow{3}{*}{TextGCN~\cite{Yao_Mao_Luo_2019}} & F1        & \underline{0.722} & \underline{0.694} & 0.671 & 0.663 & 0.730  \\
                                 && Precision & 0.715 & {0.687} & 0.677 & 0.652 & 0.675  \\
                                 && Recall    & 0.735 & {0.735} & 0.679 & {0.680} & {0.780}  \\
        \cmidrule{2-8}
        & \multirow{3}{*}{Text-Level-GNN~\cite{huang-etal-2019-text}} & F1        & 0.185 & 0.200 & 0.404 & 0.287 & 0.623  \\
                                 && Precision & 0.694 & 0.362 & 0.424 & 0.250 & 0.651  \\
                                 && Recall    & 0.183 & 0.275 & 0.424 & 0.344 & 0.606  \\
        \cmidrule{2-8}
        & \multirow{3}{*}{HyperGAT~\cite{ding-etal-2020-less}} & F1        & 0.715 & 0.372 & \underline{0.745} & 0.173 & 0.678  \\
                                 && Precision & {0.733} & 0.386 & 0.766 & 0.176 & 0.673  \\
                                 && Recall    & 0.703 & 0.374 & {0.737} & 0.178 & 0.691  \\
        \cmidrule{2-8}
        & \multirow{3}{*}{TensorGCN~\cite{Liu_You_Zhang_Wu_Lv_2020}} & F1        & 0.412 & 0.499 & 0.566 & 0.107 & \underline{0.805}  \\
                                 && Precision & 0.421 & 0.557 & 0.692 & 0.147 & {0.851}  \\
                                 && Recall    & 0.450 & 0.604 & 0.575 & 0.157 & 0.775  \\
        \cmidrule{2-8}
        & \multirow{3}{*}{SHINE~\cite{wang-etal-2021-hierarchical}} & F1        & 0.560 & 0.610 & 0.620 & 0.459 & 0.560  \\
                                 && Precision & 0.553 & 0.600 & 0.620 & 0.461 & 0.568  \\
                                 && Recall    & 0.583 & 0.623 & 0.627 & 0.473 & 0.613  \\
        \cmidrule{2-8}
        & \multirow{3}{*}{ME-GCN~\cite{wang2022megcnmultidimensionaledgeembeddedgraph}} & F1        & 0.632 & 0.603 & 0.601 & 0.607 & 0.659  \\
                                 && Precision & 0.738 & 0.601 & 0.607 & 0.629 & 0.718  \\
                                 && Recall    & 0.619 & 0.620 & 0.608 & 0.592 & 0.634  \\
        \cmidrule{2-8}
        & \multirow{3}{*}{InducT-GCN~\cite{inductgcn}} & F1        & 0.661 & 0.667 & 0.649 & 0.533 & 0.688  \\
                                 && Precision & 0.738 & 0.649 & 0.648 & 0.754 & 0.753  \\
                                 && Recall    & 0.639 & 0.711 & 0.654 & 0.478 & 0.671  \\
        \cmidrule{2-8}
        & \multirow{3}{*}{PQ-GCN} & F1        & \textbf{0.724} & \textbf{0.712} & \textbf{0.751} & {0.672} & {0.801}  \\
                                 && Precision & 0.723 & 0.695 & {0.754} & {0.692} & 0.882  \\
                                 && Recall    & {0.727} & 0.750 & 0.749 & 0.662 & 0.777 \\
        \bottomrule
    \end{tabular}
\end{table*}

We also perform an ablation study with respect to the proposed phrase and phrase POS tag features. For each experiment, the feature is excluded during embedding concatenation, and the rest of the model architecture remains as it is. The results are shown in Table~\ref{tab:ablation}.

\begin{table*}[t!]
    \centering
    \caption{Macro F1 scores obtained from ablation study}
    \label{tab:ablation}
    \begin{tabular}{lccccc}
        \toprule
        \textbf{Feature Removed} & \textbf{NU} & \textbf{ARC} & \textbf{LREC} & \textbf{Bloom} & \textbf{TREC} \\
        \midrule
        None (All Features)      & 0.724 & 0.712 & 0.751 & 0.672 & 0.801 \\
        Phrases                & 0.722 & 0.741 & 0.733 & 0.699 & 0.876  \\
        Phrase POS Tags                & 0.708 & 0.740 & 0.771 & 0.688 & 0.804  \\
        Phrases \& Phrase POS Tags                & 0.560 & 0.610 & 0.620 & 0.459 & 0.560  \\
        \bottomrule
    \end{tabular}
    
\end{table*}

\section{Analysis}

PQ-GCN achieves the best macro-averaged F1 scores across NU (0.724), ARC (0.712), and LREC (0.751), while consistently delivering high precision and recall. The improvement from the base SHINE model is also highlighted, as we have kept the original features as proposed in \cite{wang-etal-2021-hierarchical}, showing the effectiveness of phrase-based features in improving question classification capabilities.

CNN model trained in this experiment is a relatively simple model with no specialized components, yet performs well on Bloom and TREC. This can be attributed to the characteristics of the two corpus, where presence of certain words indicate a strong association to a category. An analysis of most frequently occurring words have been included in Table~\ref{tab:keywords}, showing that the Bloom and TREC datasets contains distinct and mutually exclusive class-specific keywords. For example, Bloom's \textit{Synthesis} class contains high percentages of command words such as ``suggest", ``propose", and ``design" that does not appear in questions of other classes. While the convolution kernels of CNN are effective in capturing these keywords as local features, it performs poorly on NU, ARC, and LREC where there are several overlapping non-class-specific keywords such as ``would", ``water", and ``energy".

\begin{table*}[ht!]
\centering
\caption{Keyword Analysis for Five Datasets. Frequency is calculated by \% of questions in the label set that contains the keyword.}\label{tab:keywords}
\begin{tabular}{@{}llr@{}}
\toprule
\textbf{Dataset} & \textbf{Label} & \textbf{Top Occurring Keywords (Frequency)}  \\
\midrule
\multirow{3}{*}{NU} 
          & {Transfer} & would (11.8\%), compare (6.4\%), story (6.4\%)           \\

 & {Knowledge} & words (12.5\%), describe (9.5\%), state (7.0\%)         \\

 & {Application} & construct (16\%), use (14\%), would (12.0\%)      \\
\midrule
\multirow{3}{*}{ARC} 
          & {Linguistic matching} & likely (22.1\%), best (16.0\%), water (10.7\%)      \\

 & {Basic facts} & best(16.7\%), chemical (12.8\%), new (10.3\%)     \\

 & {Hypothetical} & experiment (30.0\%), water (27.1\%), students (20.0\%)  \\

\midrule
\multirow{3}{*}{LREC} & Very short answer & energy (13.3\%), mass (11.9\%), density (11.2\%)      \\
          & Context sensitive & explain (18.6\%), think (12.1\%), lab (11.3\%)        \\
          & Answers will vary & energy (28.6\%), explain (20.8\%), would (13.0\%)      \\
\midrule
\multirow{6}{*}{Bloom} & Comprehension & explain (37.8\%), discuss (16.6\%), describe (14.3\%)      \\
          & Knowledge & list (22.4\%), define (16.0\%), five (12.8\%)        \\
          & Evaluation & justify (23.6\%), evaluate (19.2\%), answer (10.4\%)       \\
          & Application & calculate (14.6\%), explain (13.6\%), determine (11.1\%)       \\
          & Analysis & differentiate (17.8\%), compare (15.5\%), analyze (12.5\%)      \\
          & Synthesis & suggest (23.7\%), propose (11.83\%), design (10.8\%)   \\
\midrule
\multirow{6}{*}{TREC} & Entity & name (9.8\%), fear (4.9\%), first (3.9\%)      \\
          & Description and abstract concept & mean (5.23\%), origin (4.2\%), get (2.8\%)     \\
          & Human being& name (11.6\%), first (7.38\%), president (5.8\%)   \\
          & Numeric value & many (32.7\%), year (8.2\%), much (6.1\%) \\
          & Location & country (13.7\%), city (12.0\%), state (7.2\%)      \\
          & Abbreviation & stand (48.4\%), abbreviation (19.0\%), mean (11.6\%)     \\
\bottomrule
\end{tabular}
\end{table*}

PQ-GCN performs well against BERT as well, especially for NU, ARC and LREC. However, on Bloom and TREC, BERT outperforms PQ-GCN, possibly due to the difference in the size of the dataset available for fine-tuning the model.

TensorGCN performs with the highest F1 score on TREC among graph models. TensorGCN obtains its semantic embeddings by training an LSTM model on the fly during graph construction, and this method might be effective in generating more context-accurate embeddings with larger datasets. While this brings about a 0.004 point gain in F1 score over PQ-GCN, it also results in a longer graph construction time with increasing dataset size due to increased training time of the LSTM model, as compared to the proposed methodology of using pre-trained embedding models. 

Bi-LSTM has a consistently average performance across the datasets, which is expected of a simple sequence model. TextGCN shows apt results for NU and ARC datasets where its F1 scores are second only to PQ-GCN, but there is room for improvement in leveraging text features other than PMI and TF-IDF when constructing the graph, especially when it comes to text with niche characteristics such as questions. Text-Level GNN shows inconsistent poor performance across the datasets, as its learnable word-word edge weights are highly dependent on relative proximity of keywords and the ability of the message passing mechanism to disambiguate polysemous words based on word proximity. HyperGAT also shows inconsistent performance, possibly attributed to its word-word edge features, which are extracted by segmenting a text at sentence level and treating each sentence as a ``window'', as opposed to a sliding window in PMI calculation. Questions in the datasets usually consist of one or two sentences, and lacks usual word-sentence-document hierarchy that HyperGAT leverages on. ME-GCN's embedding distance-based edge features and InducT-GCN's text construction that mimics that of TextGCN also do not consider question-specific features, and PQ-GCN outperforms both models in F1 scores across all datasets.

The results of the ablation study conducted in Table~\ref{tab:ablation} shows that adding either phrase-based feature graph into training the model improves upon the base model, while their combined effectiveness varies across datasets. This variation is expected as the proposed approach extracts noun and verb phrases dynamically for each dataset using regex patterns, and the resulting phrase set and hence constructed graph differs for each dataset.

\section{Limitations}

PQ-GCN adopts a transductive approach to learning the representations, and hence it is not suited for on-line inference. While the focus of this work is to improve graph-based question classification with phrase features, we hope to inspire future work in incorporating phrase-based features into inductive methods.

We evaluate PQ-GCN only on English-based question datasets. For the proposed method to be applied to other languages, the feature extraction methods must first be adapted to syntactic rules of the target language. However, extracting regex-based noun or verb phrases may be challenging for non-configurational languages with free word order~\cite{chomsky1993lectures}.

We also evaluate PQ-GCN against BERT---a representative encoder-based language model---for a more complete comparison, but not decoder-based modern large language models, as fine-tuning these models is prohibitively expensive, and generative models are not aligned with our larger focus on evaluating additive text features in embedding-based classification.

We highlight that our method is more effective in low-resource settings compared to high-resource settings. However, labeling of a large question bank is resource-intensive for educators, and they are more likely to have a smaller subset of questions to be analyzed, in which case our proposed method would be succinct.

\section{Conclusion}

We develop and evaluate PQ-GCN, which augments a SHINE-based GCN for question classification by incorporating phrase features into the text graphs that can provide deeper semantic connections within questions. The proposed method shows competitive classification performance compared to baseline models, especially in low-resource settings without the need for extensive pre-training. Question classification is an important task in building educational analytics systems, as we move towards an era of personalized learning with artificial intelligence. Finding suitable feature extraction methods is crucial in tackling downstream tasks in an increasingly data-scarce landscape, and our proposed method provides a parameter- and data-efficient approach to enhancing question classification.

\bibliography{mybibfile}

\end{document}